\documentclass[conference]{IEEEtran}
\usepackage{times}

\usepackage[numbers]{natbib}
\usepackage{multicol}
\usepackage[bookmarks=true]{hyperref}
\usepackage{booktabs}
\usepackage{multirow}
\usepackage{array}
\usepackage{amsmath}
\usepackage{amssymb}
\usepackage{graphicx}
\usepackage{dblfloatfix}
\usepackage{caption}
\usepackage[table]{xcolor}

\begin{document}

% paper title
\title{UniVTAC: A Unified Simulation Platform for Visuo-Tactile Manipulation Data Generation, Learning, and Benchmarking}

% You will get a Paper-ID when submitting a pdf file to the conference system
% \author{Author Names Omitted for Anonymous Review. Paper-ID [xxx]}

\author{\authorblockN{
Baijun Chen$^{5,2*}$,
Weijie Wan$^{6*}$,
Tianxing Chen$^{4*}$, 
Xianda Guo$^{7,2*}$,
Congsheng Xu$^{1}$,\\
Yuanyang Qi$^{3}$,
Haojie Zhang$^{3}$,
Longyan Wu$^{8}$,
Tianling Xu$^{1}$,
Zixuan Li$^{6}$, 
Yizhe Wu$^{3}$,\\
Rui Li$^{3,9}$,
Xiaokang Yang$^{1}$,
Ping Luo$^{4}$,
Wei Sui$^{2,\dagger}$, and
Yao Mu$^{1,\dagger}$}\\

$^{1}$ ScaleLab, Shanghai Jiao Tong University
$^{2}$ D-Robotics
$^{3}$ ViTai Robotics
$^{4}$ The University of Hong Kong\\
$^{5}$ Nanjing University
$^{6}$ Shenzhen University
$^{7}$ Wuhan University
$^{8}$ Fudan University
$^{9}$ Tsinghua University \\

$^*$Equal Contribution  $^\dagger$Corresponding Authors
}
\maketitle

\begin{abstract}
Robotic manipulation has seen rapid progress with vision-language-action (VLA) policies. However, visuo-tactile perception is critical for contact-rich manipulation, as tasks such as insertion are difficult to complete robustly using vision alone. At the same time, acquiring large-scale and reliable tactile data in the physical world remains costly and challenging, and the lack of a unified evaluation platform further limits policy learning and systematic analysis. To address these challenges, we propose UniVTAC, a simulation-based visuo-tactile data synthesis platform that supports three commonly used visuo-tactile sensors and enables scalable and controllable generation of informative contact interactions. Based on this platform, we introduce the UniVTAC Encoder, a visuo-tactile encoder trained on large-scale simulation-synthesized data with designed supervisory signals, providing tactile-centric visuo-tactile representations for downstream manipulation tasks. In addition, we present the UniVTAC Benchmark, which consists of eight representative visuo-tactile manipulation tasks for evaluating tactile-driven policies. Experimental results show that integrating the UniVTAC Encoder improves average success rates by 17.1\% on the UniVTAC Benchmark, while real-world robotic experiments further demonstrate a 25\% improvement in task success. Our webpage is available at \href{https://univtac.github.io/}{https://univtac.github.io/}.
\end{abstract}

\IEEEpeerreviewmaketitle

\section{Introduction}
\label{sec:intro}

Recent advances in robotic manipulation have enabled impressive performance across a broad range of tasks. Modern policies~\cite{intelligence2025pi,rdt2,graspvla,wen2025dexvla,liu2025hybridvla,team2024octo,li2024cogact,kim2024openvla,black2024pi_0,liu2024rdt1b,chen2025g3flow,fu2025cordvip} demonstrate strong capabilities in flexible object manipulation and fine-grained skills, supporting complex activities such as coffee making and tool use. Nevertheless, precise and reliable manipulation remains fundamentally challenging, particularly for tasks that require accurate reasoning about object pose and contact states, including insertion and alignment. In these settings, vision-based perception alone is often insufficient due to occlusions introduced by the end-effector, limited depth accuracy at close range, and the lack of direct observability of contact interactions once physical engagement occurs.

Tactile sensing provides complementary information by directly capturing localized contact geometry, force distribution, and relative motion at the interaction interface, which enables the detection of misalignment and supports closed-loop corrective control during manipulation. Building upon these advantages, visuo-tactile perception integrates visual appearance with tactile deformation cues, yielding dense and spatially resolved observations of contact phenomena such as pressure distribution, surface texture, and material properties. This combination makes visuo-tactile perception particularly well suited for contact-rich manipulation scenarios where reliable interaction depends on accurate perception under occlusion and close-range contact.

However, the current infrastructure for visuo-tactile manipulation remains underdeveloped. On the one hand, the scarcity of large-scale tactile data severely limits the training of tactile-centric representation models, leading to suboptimal performance of tactile-based manipulation policies. This limitation largely stems from practical constraints of real-world tactile sensing hardware, including the lack of standardized sensor designs, high manufacturing and deployment costs, and the difficulty of large-scale production and data collection. As a result, both representation learning and policy optimization for tactile manipulation are significantly constrained.

On the other hand, the absence of unified and comprehensive benchmarks for visuo-tactile manipulation hinders systematic evaluation and iterative improvement of tactile-driven strategies. Addressing this gap requires a reproducible platform that supports both scalable data acquisition and standardized policy evaluation, enabling fair comparison and principled analysis across different methods.

To address these challenges, we propose UniVTAC, a simulation-based framework for synthesizing visuo-tactile manipulation data that supports three widely used visuo-tactile sensors. Leveraging high-fidelity simulation, UniVTAC enables large-scale and controllable generation of visuo-tactile interaction data, including pressure patterns, marker deformations, and tangential force cues. Based on the synthesized data, we design multiple supervision objectives, such as visuo-tactile image reconstruction and pose estimation, and train a unified visuo-tactile encoder under this multi-task supervision. Through large-scale pretraining, the encoder learns tactile-centric representations that capture fine-grained contact boundaries while remaining sensitive to object pose and interaction dynamics.

To systematically evaluate the effectiveness of the synthesized data and the learned representations, we further introduce a simulation-based visuo-tactile manipulation benchmark built on TacEx~\cite{nguyenTacExGelSightTactile2024} and NVIDIA Isaac Sim~\cite{NVIDIA_Isaac_Sim}. The benchmark comprises eight representative tactile manipulation tasks and supports automated data synthesis and unified policy evaluation. Together with UniVTAC, it enables systematic and reproducible analysis of visuo-tactile manipulation strategies under diverse contact scenarios.
Our contributions are summarized as follows:

(1) \textbf{Simulation-based visuo-tactile data synthesis.}
We propose UniVTAC, a scalable and controllable simulation framework for synthesizing visuo-tactile contact and manipulation data, supporting three commonly used visuo-tactile sensors. Within this framework, we design task-specific supervisory signals tailored for tactile manipulation and use them to generate large-scale annotated synthetic visuo-tactile data. Leveraging this data, we pretrain a visuo-tactile encoder designed to support downstream tactile manipulation tasks.

(2) \textbf{UniVTAC Benchmark.}
We introduce the UniVTAC Benchmark, a simulation-based benchmark comprising eight dexterous visuo-tactile manipulation tasks. Built upon the UniVTAC platform, the benchmark supports automated data generation and unified policy evaluation, enabling systematic and reproducible analysis of visuo-tactile manipulation strategies.

(3) \textbf{Comprehensive evaluation and real-world validation.}
We conduct extensive experiments the on UniVTAC benchmark to analyze the performance of representative manipulation policies on tactile-dependent tasks. We demonstrate the effectiveness of the UniVTAC Encoder representations in simulation and further validate their applicability through real-world robotic experiments.
\section{Related Work}
\label{sec:related_work}

\subsection{Visuo-tactile Sensor Simulation}
The development of high-fidelity simulators for visuo-tactile sensors is a cornerstone for data-driven tactile perception research. Recent progress has been driven by advancements in physics-based simulation and differentiable rendering. Existing frameworks can be categorized by their underlying physical formulation.

The Incremental Potential Contact (IPC) method \cite{li2020incremental} has emerged as a prominent framework for contact-rich robotic simulation, offering rigorous non-penetration guarantees and robust handling of complex frictional interactions. Building on IPC, \cite{chenGeneralPurposeSim2RealProtocol2024} and \cite{liManiSkillViTac2025Challenge2024} integrated it into robotic manipulation pipelines through SapienIPC, a soft-body capable extension of the original SAPIEN simulator, enabling high-fidelity sim-to-real transfer. Taccel~\cite{liTaccelScalingVisionbased2025} significantly accelerated this pipeline via a GPU-optimized IPC backend, achieving real-time or faster-than-real-time simulation. TacEx~\cite{nguyenTacExGelSightTactile2024} couples the modern IPC library libuipc~\cite{gipc2024,stiffgipc2025} with a physically based visuo-tactile renderer in NVIDIA Isaac Sim, supporting real-time, high-fidelity simulation of GelSight-style sensors. The Finite Element Method (FEM) offers an alternative route focused on geometric and material fidelity, exemplified by TacFlex~\cite{zhangTacFlexMultimodeTactile2025}. By coupling FEM-based elastic deformation models with ray-traced optical rendering. In contrast, DiffTactile~\cite{si2024difftactile} adopts the Material Point Method (MPM) to model highly deformable objects, leveraging its compatibility with automatic differentiation for gradient-based policy optimization. Separately, Tacchi~\cite{chenTacchiPluggableLow2023} utilizes MPM for its efficiency in pluggable, modular simulation settings, where fast setup and moderate accuracy are prioritized.

Our work is built upon the TacEx framework~\cite{nguyenTacExGelSightTactile2024}, leveraging its physical accuracy and integration in Isaac Sim, introducing multiple sensors support and automatic manipulation APIs. Our marker deformation simulation is also inspired by the IPC-based contact modeling in \cite{chenGeneralPurposeSim2RealProtocol2024} and \cite{liManiSkillViTac2025Challenge2024}.

\subsection{Visuo-tactile Representation Learning}

Learning effective representations from tactile signals typically follows three paradigms: reconstruction, explicit geometric supervision, and multi-modal alignment. Reconstruction-based methods focus on capturing visual distribution. Marker-Embedded GAN \cite{kimMarkerEmbeddedTactileImage2023} utilizes GANs to generate RGB images from depth maps to bridge the cross-modal gap, while UniT \cite{xuUniTDataEfficient2025} employs VQGAN to reconstruct tactile images for data-efficient encoding. However, as these methods optimize for pixel-level fidelity, they often overlook the underlying contact physics and force distributions.

To capture finer contact details, explicit geometric supervision has been explored. RDP~\cite{xueReactiveDiffusionPolicy2025} treats markers as a 2D grid-structured point matrix and applies PCA to capture dominant deformation modes. In contrast, the framework of \cite{chenGeneralPurposeSim2RealProtocol2024} models markers as 3D point clouds and uses a PointNet-based architecture to learn nonlinear geometric features from full spatial coordinates. While these methods are robust in modeling physical contact, they often sacrifice rich tactile textures and object-specific shape information by discarding high-dimensional image features.

Recently, vision-tactile contrastive learning has gained traction by aligning tactile embeddings with multi-view or temporal visual signals \cite{georgeVITaLPretrainingVisuoTactile2024, liuViTaMInLearningContactRich2025}. Although these approaches achieve impressive results in contact-rich manipulation, the resulting representations are often task-specific. Relying on global alignment (e.g., CLIP-like objectives) often necessitates training the encoder for each new task, thereby limiting the generalizability of the learned tactile representation.

\subsection{Simulation for Data Generation}

Many simulation-based methods have been proposed for robotic policy learning~\cite{xiang2020sapien,maniskill2,metaworld,mees2022calvin,robocasa2024,geng2025roboverse,garcia24gembench,pumacay2024colosseum,liu2023libero,lan2025autobio,chen2025robotwin,mu2025robotwin}, providing scalable environments for training and evaluation. Building upon these platforms, a number of simulation-based pipelines further focus on large-scale robot data generation. For example, the RoboTwin series~\cite{mu2024robotwin, mu2025robotwin, chen2025robotwin} and RoboGen~\cite{wang2023robogen} employ expert code to synthesize high-quality, near-perfect manipulation trajectories. MimicGen~\cite{mandlekar2023mimicgen} and DexMimicGen~\cite{jiang2024dexmimicen} extend data collection through human teleoperation, while other works~\cite{yu2020meta,li2025simplevla} obtain expert policies via reinforcement learning with carefully designed reward functions and subsequently use them for data synthesis. Collectively, these approaches enable efficient large-scale data collection and, in some cases, support natural-language task specifications for novel tasks~\cite{liu2025hycodepolicy}.

Despite their success, most existing simulation-based data generation pipelines primarily focus on rigid or articulated object manipulation and provide limited support for modeling the complex dynamics of tactile interaction. Standard protocols typically rely on heuristic gripper commands and binary success criteria based on final object states, without explicitly modeling transient contact forces, deformation, or slippage during execution. As a result, such open-loop or weakly constrained execution may produce trajectories with excessive grasping forces or unstable contact configurations, which would be unsafe for real-world tactile sensors and insufficient for learning tactile-dependent manipulation skills.

In contrast, our work explicitly incorporates tactile awareness into the data generation process through a closed-loop control scheme for gripper aperture. By integrating runtime validity checks and contact-sensitive feedback, our pipeline avoids sensor-destructive behaviors and ensures that synthesized trajectories capture rich, physically meaningful contact responses. This design enables the generation of tactile-aware data that is better aligned with the requirements of visuo-tactile manipulation and downstream tactile-based policy learning.
\section{UniVTAC}
The representation of observations plays a central role in effective robotic manipulation, particularly in contact-rich settings. However, acquiring large-scale and reliable tactile data in the physical world remains difficult to scale, as real-world tactile sensors provide only raw observations and incur substantial data collection costs. Simulation environments, by contrast, offer access to ground-truth physical states and enable scalable synthesis of contact interactions.

In this section, we present UniVTAC as an end-to-end simulation-driven framework for visuo-tactile manipulation, encompassing data synthesis, representation learning, and systematic evaluation. UniVTAC enables automated generation of large-scale annotated visuo-tactile interaction data, which we use to train the UniVTAC Encoder via auxiliary supervision derived from simulation-specific physical signals. The resulting tactile-centric visuo-tactile representations are designed to support downstream policy learning in contact-rich manipulation tasks. To systematically assess the effectiveness of the learned representations and the policies built upon them, we further introduce the UniVTAC Benchmark, a simulation benchmark comprising eight representative visuo-tactile manipulation tasks that span diverse contact scenarios.
\subsection{UniVTAC Platform}
Built upon the TacEx framework \cite{nguyenTacExGelSightTactile2024}, UniVTAC extends the soft-body simulation capabilities to support diverse sensor types and complex manipulation tasks.

\subsubsection{Sensor Configuration and Heterogeneity}

UniVTAC integrates three mainstream visuotactile sensors: \textit{GelSight Mini}\cite{yuan2017gelsight}, \textit{ViTai GF225}~\cite{vitai_sdk_release}, and \textit{Xense WS}~\cite{Xense2025}. The optical and mechanical properties of each sensor are modeled by adjusting internal camera intrinsics, gelpad meshes, and rendering methods. This modularity allows researchers to validate tactile algorithms across different hardware specifications within a unified environment.

\subsubsection{Automated Manipulation APIs}

To facilitate large-scale and high-fidelity data collection, we implement a library of atomic manipulation APIs: \textit{Grasp}, \textit{Move}, \textit{Place}, \textit{Probe}, and \textit{Rotate}. While \textit{Move} and \textit{Place} focus on precise trajectory generation via \textit{cuRobo} \cite{sundaralingamCuRoboParallelizedCollisionFree2023}, the \textit{Grasp} and \textit{Probe} primitives incorporate a tactile-reactive adaptive control mechanism to ensure physical consistency. Specifically, during the \textit{Grasp} sequence, the gripper's joint velocity $\dot{q}$ is governed by a state-dependent feedback law to prevent non-physical penetration:

\begin{equation}
\dot{q} =
\begin{cases}
    v_{\text{fast}} & \text{if } d_{\min} = d_{\max} \\
    \min \left(|d_{\min} - \delta_{\text{th}}|, v_{\text{slow}} \right) & \text{if } d_{\min} < d_{\max}
\end{cases}
\end{equation}

where $d_{\min}$ is the real-time minimum depth from the tactile sensors, $d_{\max}$ denotes the zero-contact depth, and $\delta_{\text{th}}$ is the reaching depth threshold. The $v_{\text{fast}}$ and $v_{\text{slow}}$ constrains the maximum velocity during the phase without or with contact. This closed-loop approach effectively mitigates non-physical clipping artifacts and ensures that the captured tactile imprints remain within a realistic deformation manifold, providing high-quality training data for the subsequent representation learning.

\subsubsection{Tactile Representation Prerequisites}

\begin{figure*}[t]
    \centering
    \includegraphics[width=0.95\textwidth]{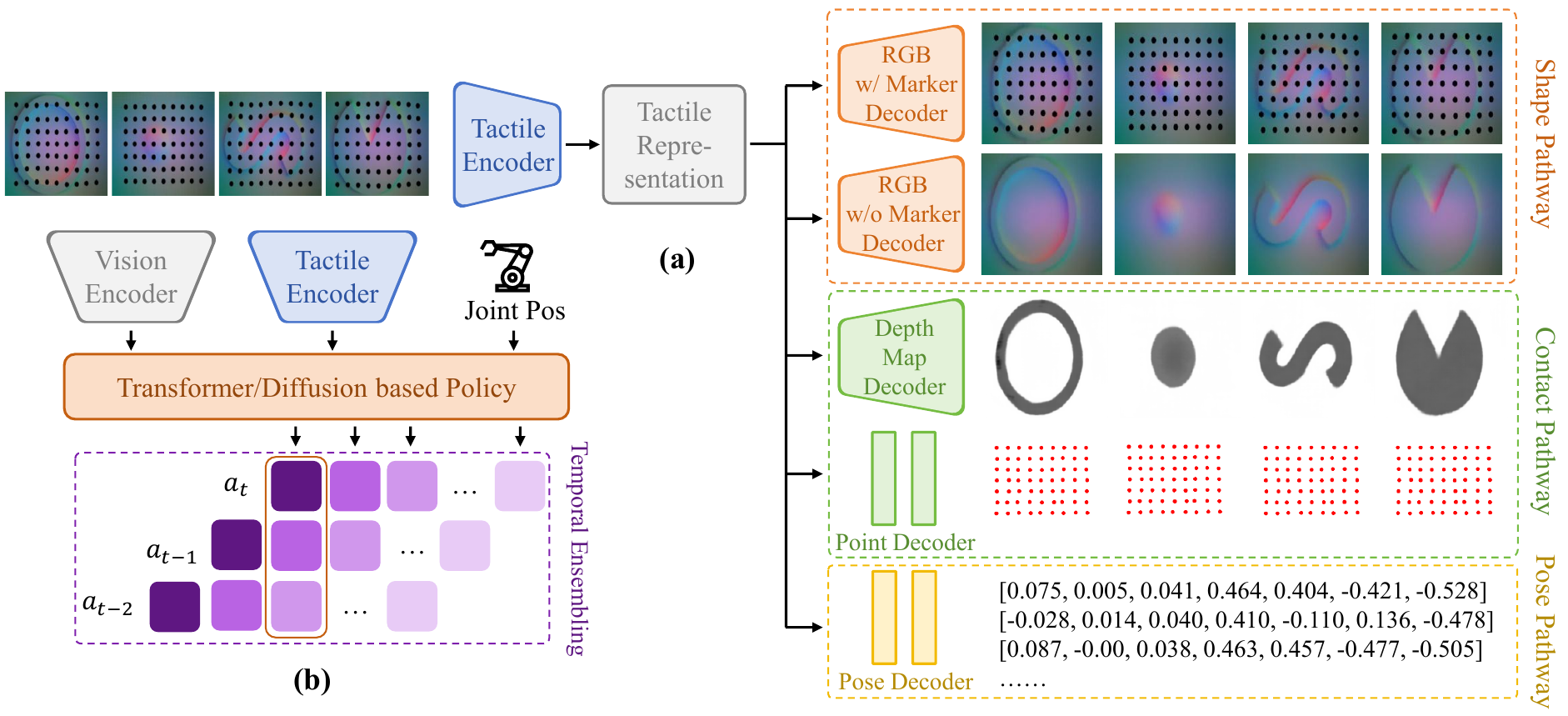}
    \caption{\textbf{The UniVTAC Encoder Framework and Integration into Policy Learning.} (a) The UniVTAC Encoder is pretrained with three self-supervised objectives, including shape reconstruction, contact deformation prediction, and object pose regression, to learn a structured tactile-centric representation from raw visuo-tactile observations. (b) At deployment time, the pretrained encoder is integrated as a perception module for downstream manipulation policies, enabling end-to-end policy learning from raw tactile images without introducing additional inference-time overhead.}
    \label{fig:encoder}
\end{figure*}

To learn a robust general-purpose tactile representation, we posit that the encoder must disentangle three distinct physical properties from raw tactile observations. We categorize these into three perception pathways, which guide our data generation and subsequent supervision strategies. \textbf{Shape Perception} targets global geometry, compelling the model to disentangle the object's intrinsic visual appearance from sensor-specific artifacts such as markers and lighting. \textbf{Contact Perception} focuses on local dynamics, explicitly modeling interaction mechanics through surface deformation and shear-induced marker displacement. \textbf{Pose Perception} provides spatial grounding, anchoring tactile signals in external metric space to enable consistent spatial reasoning for downstream policies. To support these pathways, our pipeline is designed to capture high-fidelity ground truth for each modality alongside the raw sensor input.

\subsubsection{Contact-Rich Data Generation}

We curate a geometrically diverse dataset to ensure the encoder encounters broad variations in local curvature and contact topology. Following protocols in \cite{kimMarkerEmbeddedTactileImage2023}, we employ 14 distinct geometric primitives as indenters, ranging from standard convex shapes (e.g., \textit{spheres, cones}) to complex non-convex patterns (e.g., \textit{stars, cross-shapes}). These objects are mounted on a standardized prism base to facilitate varying grasp interactions.

For each episode, the robotic gripper approaches a prism on the ground plane with shapes on its surface and initiates contact. To capture the dynamic relationship between gripper width and gelpad deformation, we vary the grasping tightness stochastically by setting the reaching depth threshold $\delta_{\text{th}}$. This generates data ranging from light touches to deep indentations. Once in contact, the robot executes a series of randomized small-scale rotations through the \textit{Move} and \textit{Rotate} APIs. These actions induce rich shear force patterns and marker displacements, which are critical for the contact perception. During this interaction sequence, we synchronously record the raw tactile images with markers $I_{\text{marked}}$, along with the simulation-privileged ground truth: pure contact images $I_{\text{pure}}$, depth maps $D$, 2D projections of fiducial markers $M$, and the pose of the object in the gelpad center local frame $p$.

This automated pipeline generates approximately 14,000 interaction frames per shape, yielding a total dataset of 205,826 samples.

\subsection{UniVTAC Encoder}

We formulate the UniVTAC Encoder as a multi-pathway representation learning framework that embeds structured physical priors into tactile observations through task-driven supervision. As illustrated in Figure~\ref{fig:encoder}, a shared encoder maps each visuo-tactile observation into a compact latent representation, which is then decoded by multiple pathway-specific heads during training. These decoding pathways impose complementary inductive biases on the learned representation, while all decoders are discarded at deployment time, ensuring that only the encoder is retained with no additional inference-time overhead.

\begin{figure*}[h]
    \centering
    \includegraphics[width=1.0\textwidth]{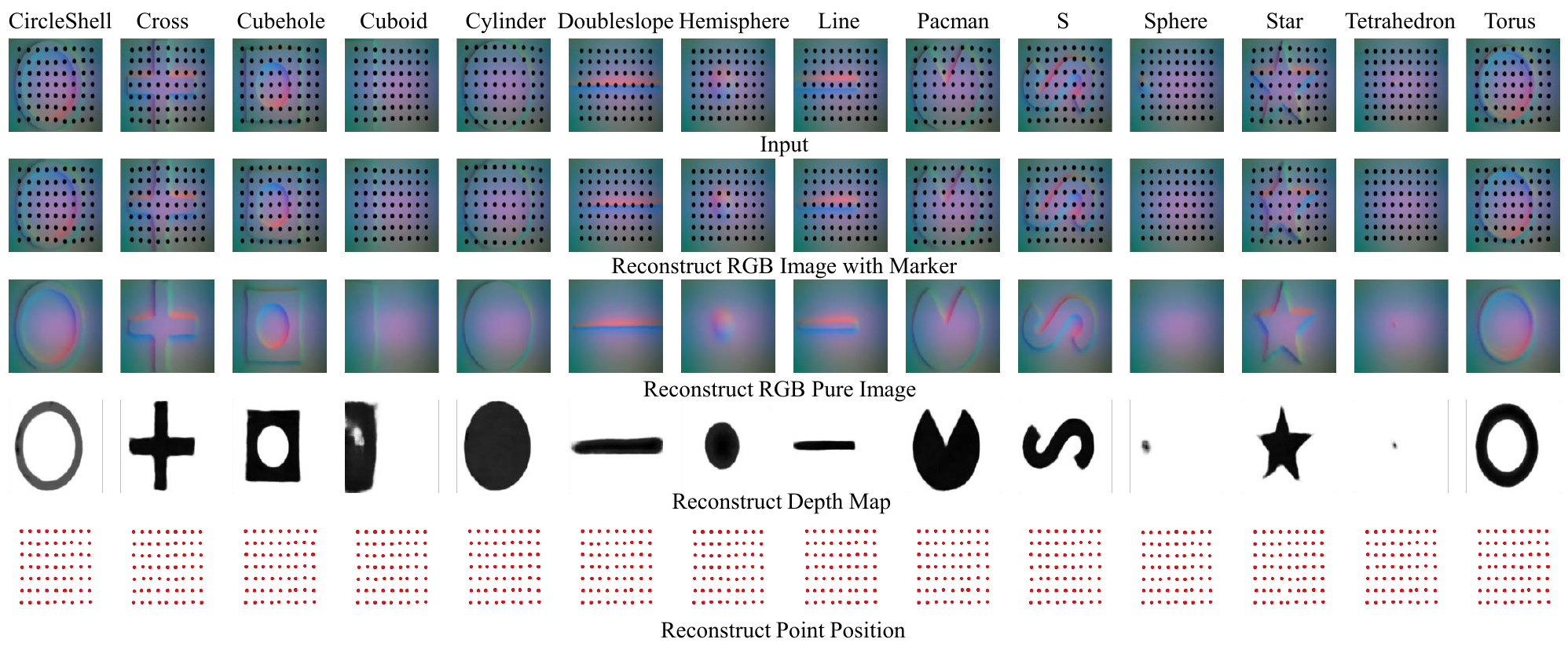}
    \caption{\textbf{Reconstruction Result}. From a tactile image with markers, the UniVTAC Encoder reconstructs complementary physical signals, including the marker-free tactile image, gelpad deformation depth map, and marker point positions, across diverse contact geometries. These results show that the learned representation captures both global shape cues and fine-grained contact deformation beyond sensor-specific visual patterns.}
    \label{fig:reconstruction}
\end{figure*}

The design of UniVTAC is grounded in three perceptual prerequisites that are critical for contact-rich manipulation: object shape understanding, local contact dynamics, and spatial pose awareness. To support shape perception, the UniVTAC Encoder employs a dual-view reconstruction strategy in which the encoder is supervised to reconstruct both the raw tactile image with marker patterns, denoted as $I_{\text{marked}} \in \mathbb{R}^{3 \times H \times W}$, and a counterfactual marker-free image $I_{\text{pure}} \in \mathbb{R}^{3 \times H \times W}$. Jointly optimizing these two reconstruction objectives encourages the encoder to disentangle intrinsic object geometry from sensor-specific marker artifacts, resulting in shape-aware representations that are robust to appearance variations.

Contact perception is enforced by supervising the encoder with physically grounded signals that capture local deformation dynamics. Specifically, the UniVTAC Encoder predicts a dense surface depth map $D \in \mathbb{R}^{H \times W}$ representing normal indentation of the gelpad, together with the 2D projections of $N$ fiducial markers, denoted as $M \in \mathbb{R}^{2 \times N}$, which encode lateral shear and tangential deformation. By jointly modeling geometric deformation and marker displacement, the learned representation is encouraged to capture true contact mechanics rather than relying on pixel-level cues alone.

To anchor tactile observations within a global spatial context, the UniVTAC Encoder further incorporates a pose perception pathway that regresses the relative object pose $p \in \mathbb{R}^{7}$, consisting of 3D translation and a quaternion representation of orientation. This explicit spatial supervision grounds the latent representation in a metric space, enabling downstream policies to perform spatially consistent reasoning during manipulation.

Architecturally, the UniVTAC Encoder adopts a ResNet-18 backbone as the shared encoder, which projects each tactile observation into a latent feature vector. The shape reconstruction heads employ transposed convolutional decoders to generate RGB outputs, while the depth decoder follows a similar architecture with a single-channel output. Marker flow is predicted using a lightweight multilayer perceptron, and pose regression is performed by a compact MLP that outputs the 7-dimensional pose vector. Reconstruction examples are shown in Figure ~\ref{fig:reconstruction}.

The shared encoder and task-specific heads are trained end-to-end using a multi-task objective. We employ Mean Squared Error (MSE) as the primary optimization criterion for all pathways. The total loss $\mathcal{L}_{\text{total}}$ is defined as a weighted sum of the following components:

\textbf{Shape Reconstruction Loss.} This loss encourages the model to capture global geometry by reconstructing both the marked and marker-free observations:

\begin{equation}
\mathcal{L}_{\text{shape}} = MSE(\hat{I}_{\text{marked}}, I_{\text{marked}}) + MSE(\hat{I}_{\text{pure}}, I_{\text{pure}})
\end{equation}
where $\hat{I}$ and $I$ denote the predicted and ground-truth tactile images, respectively.

\textbf{Contact Deformation Loss.} To supervise local interaction dynamics, we minimize the discrepancy in surface depth and marker displacements:
\begin{equation}
\mathcal{L}_{\text{contact}} = MSE(\hat{D} , D) + MSE(\hat{M},M)
\end{equation}
where $D$ is the depth map and $M$ represents the marker positions porjected onto the image plane.

\textbf{Pose Regression Loss.} The spatial grounding is optimized via:
\begin{equation}
\mathcal{L}_{\text{pose}} = MSE(\hat{p}, p)
\end{equation}
where $p$ represents the object relevant pose (3D translation and 4D quaternion).

The final training objective is
\begin{equation}
\mathcal{L}_{\text{total}} = \lambda_s \mathcal{L}_{\text{shape}} + \lambda_c \mathcal{L}_{\text{contact}} + \lambda_p \mathcal{L}_{\text{pose}}
\end{equation}

In our experiments, we empirically set the balancing hyperparameters to $\lambda_s=1.0$, $\lambda_c = 0.5$, and $\lambda_p = 0.5$.

\subsection{UniVTAC Benchmark}
\begin{figure*}[t]
    \centering
    \includegraphics[width=1.0\textwidth]{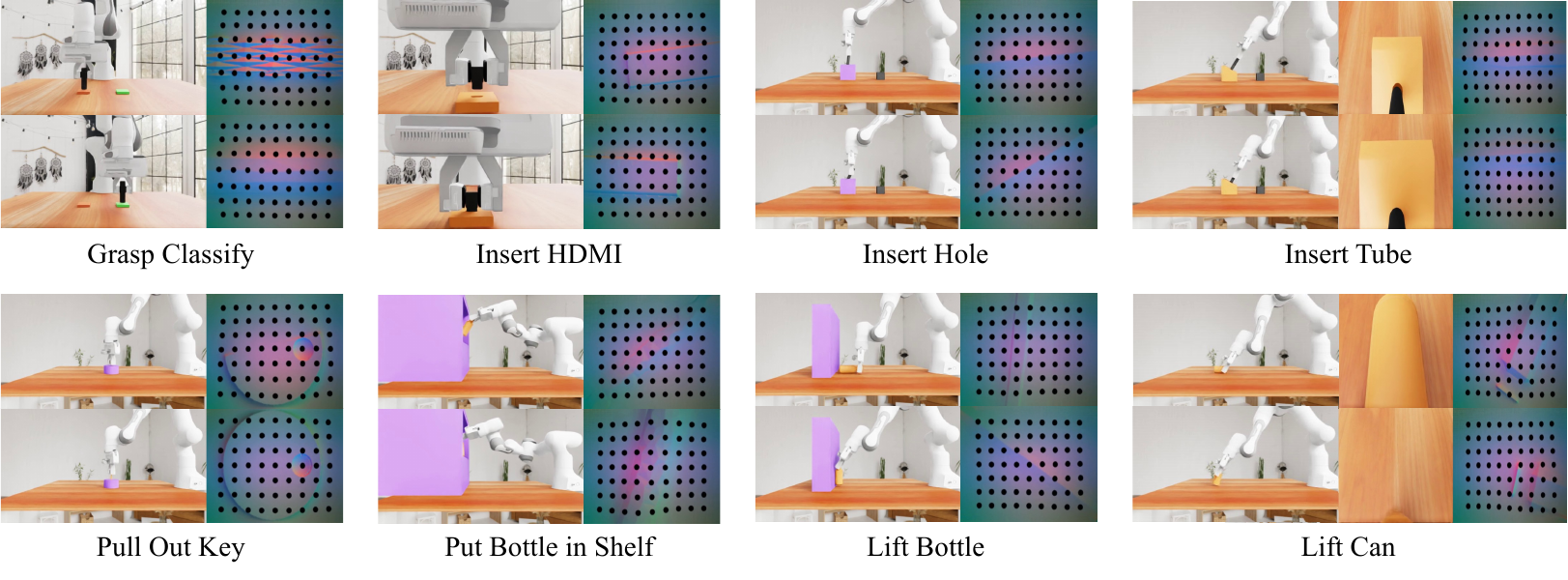}
    \caption{\textbf{UniVTAC Benchmark Tasks}. The UniVTAC Benchmark comprises eight representative visuo-tactile manipulation tasks spanning shape recognition, pose reasoning, and contact-rich interaction, and is designed to systematically evaluate tactile-dependent manipulation policies. For each task, we visualize two representative key frames corresponding to critical stages of execution. Each key frame includes both a visuo-tactile observation and a standard visual observation. For clarity of presentation, we display the tactile observation from only one side of the gripper, although tactile sensing is available on both fingertips during execution.}
    \label{fig:tasks}
\end{figure*}

Following the RoboTwin paradigm and the tactile perception design of UniVTAC, we leverage simulation APIs, annotated object assets, and expert programs to enable automated task-level data synthesis and evaluation. This design integrates data collection, model training, and policy evaluation into a unified pipeline, supporting scalable and reproducible experimentation.

Expert-driven data synthesis methods are capable of generating near-perfect manipulation trajectories. For example, insertion tasks can be synthesized with highly precise and collision-free executions. However, such trajectories are suboptimal for tactile learning, as they lack meaningful variations in contact perception. In these executions, contact events are often trivial or instantaneous, preventing tactile policies from learning informative, contact-dependent behaviors.

To address this limitation, we introduce stochasticity and corrective behaviors into the synthesis process. Specifically, for contact-rich skills such as insertion, we intentionally introduce randomized failures during execution and allow the expert controller to perform corrective actions based on perceived contact feedback, which in simulation is provided by ground-truth physical states. This process more closely resembles human manipulation behavior, where coarse alignment is followed by iterative contact-based correction until task completion. As a result, the synthesized trajectories exhibit diverse contact patterns and corrective interactions, providing rich tactile supervision that enables policies to effectively leverage visuo-tactile information.

To ensure reliable evaluation, the UniVTAC Benchmark incorporates tactile-specific, physics-based success criteria that extend beyond reaching target coordinates. A trial is considered invalid if the maximum penetration depth exceeds a predefined safety threshold or if significant relative slip is detected between the gelpad and the object surface. These constraints prevent degenerate solutions that exploit simulation artifacts and ensure that evaluation outcomes reflect physically meaningful manipulation behaviors rather than numerical or physical loopholes.

We design eight visuo-tactile manipulation tasks spanning three perceptual dimensions. All tasks are visualized in Figure~\ref{fig:tasks}, with detailed task descriptions provided in Appendix~\ref{task-description}. Pose reasoning tasks, including \textit{Lift Bottle}, \textit{Lift Can}, and \textit{Put Bottle in Shelf}, require policies to estimate the relative spatial relationship between the gripper and the object during interaction, relying primarily on accurate perception of object position and orientation. Shape perception tasks, such as \textit{Grasp Classify}, focus on distinguishing object geometries from high-dimensional tactile observations, emphasizing the ability of learned representations to capture shape-related contact cues. Contact-rich interaction tasks, including \textit{Insert Hole}, \textit{Insert Tube}, \textit{Insert HDMI}, and \textit{Pull Out Key}, involve fine-grained alignment and sequential contact transitions, requiring precise reasoning over contact dynamics throughout the manipulation process.
\section{Experiment}
\label{sec:experiment}

Our experiments are designed to investigate three key questions. (1) The performance of representative manipulation policies on the UniVTAC Benchmark are evaluated to characterize their behavior across diverse visuo-tactile tasks; (2) The efficacy of the UniVTAC Encoder are examed, comparing with existing approaches and analyzing the contributions of different pathways; (3) We conduct simulation-to-real experiments to assess whether the UniVTAC Encoder, trained solely on simulation-synthesized data, transfers effectively to real-world robotic manipulation.

\begin{table*}[!t]
\centering
\caption{\textbf{UniVTAC Benchmark}. We report the success rates and average performance of ACT without tactile input, VITaL, and ACT with the UniVTAC Encoder (Ours) across the eight tasks in the UniVTAC Benchmark. (\textbf{Bold}: the best results; \underline{Underlined}: the second-best results)}
\label{tab:benchmark}
\small
\setlength{\tabcolsep}{3.2pt}
\renewcommand{\arraystretch}{1.0}
\resizebox{\textwidth}{!}{%
\begin{tabular}{l|cccccccc|c}
\toprule
\rowcolor{gray!10}
\textbf{Method} & \textbf{Lift Bottle} & \textbf{Pull-out Key} & \textbf{Lift Can} & \textbf{Put Bottle in Shelf} & \textbf{Insert Hole} & \textbf{Insert HDMI} & \textbf{Insert Tube} & \textbf{Grasp Classify} & \textit{\textbf{Average}} \\
\midrule
ACT~\cite{zhaoLearningFineGrainedBimanual2023} & 42.0 & 28.0 & \underline{20.0} & 28.0 & 19.0 & \underline{15.0} & \underline{45.0} & 50.0  & 30.9 \\
VITaL~\cite{georgeVITaLPretrainingVisuoTactile2024}           & \textbf{72.0} & \textbf{47.0} & 8.0  & \textbf{32.0} & \textbf{25.0} & 6.0  & 34.0 & \textbf{100.0} & \underline{40.5} \\
\rowcolor{cyan!10}
\textbf{Ours}   & \underline{71.0} & \underline{46.0} & \textbf{29.0} & \underline{31.0} & \underline{24.0} & \textbf{28.0} & \textbf{56.0} & \underline{99.0}  & \textbf{48.0} \\
\bottomrule
\end{tabular}%
}
\end{table*}

\subsection{Evaluation on the UniVTAC Benchmark}
\label{subsec:benchmark}

We evaluate representative manipulation policies on the UniVTAC Benchmark, focusing on Action Chunking Transformers (ACT)~\cite{zhaoLearningFineGrainedBimanual2023}. ACT models manipulation as sequence prediction using a Transformer-based conditional variational framework, enabling temporally coherent action generation and flexible fusion of sensory inputs through attention mechanisms. To assess the effectiveness of tactile-centric representations, we further evaluate ACT augmented with the UniVTAC Encoder, which incorporates visuo-tactile features learned through large-scale simulation pretraining. In addition, we include VITaL~\cite{georgeVITaLPretrainingVisuoTactile2024}, a representative visuo-tactile manipulation policy that leverages visuo-tactile pretraining, serving as a strong baseline that explicitly exploits tactile representations.

All policies are trained on 50 automatically collected full trajectories per task and evaluated over 100 test rollouts.

Table~\ref{tab:benchmark} summarizes the performance of representative manipulation policies on the UniVTAC Benchmark. Overall, the results highlight both the effectiveness of the benchmark design and the importance of tactile feedback for contact-rich manipulation. The benchmark spans tasks with varying degrees of perceptual and interaction complexity, enabling a nuanced analysis of how different policies exploit visual and tactile information.

Across all tasks, ACT augmented with the UniVTAC Encoder achieves a substantial performance improvement over its vision-only counterpart, increasing the average success rate from 30.9\% to 48.0\%. The improvement is consistently observed across multiple contact-sensitive tasks, including insertion and pull-out scenarios. This trend indicates that the UniVTAC Encoder provides tactile-centric representations that can be effectively exploited by attention-based policy architectures. Moreover, these results suggest that the UniVTAC Benchmark captures manipulation scenarios in which tactile perception plays a decisive role, and that ACT’s Transformer-based attention mechanism is well suited for integrating heterogeneous visuo-tactile features.

We also observe that VITaL achieves strong performance on several tasks, most notably attaining near-perfect accuracy on \textit{Grasp Classify}. This behavior is expected, as VITaL benefits from explicit visuo-tactile pretraining, enabling the extraction of discriminative tactile features that are closely aligned with shape-related contact cues. As such, VITaL serves as a meaningful reference point, illustrating the effectiveness of representation pretraining for visuo-tactile manipulation and providing a practical upper bound for evaluating the UniVTAC Encoder under comparable experimental settings.

These observations indicate that the UniVTAC Benchmark spans a continuum of task dependencies, ranging from visually solvable manipulation to scenarios that fundamentally rely on tactile feedback. While vision-only policies can achieve competitive performance on tasks dominated by geometric visibility, tactile perception becomes essential for robust execution in contact-rich and correction-intensive settings. Consequently, the benchmark enables a more principled evaluation of manipulation policies by revealing not only aggregate task success, but also how effectively different policy architectures exploit visuo-tactile representations.

\subsection{Efficacy of the UniVTAC Encoder}
\label{subsec:UniVTAC Encoder_efficacy}

We further evaluate the UniVTAC Encoder by comparing it against baselines and ablation variants.

We first compare our method with a CLIP-based visuo-tactile encoder. As shown in Table \ref{tab:benchmark}, when integrated with ACT, the UniVTAC Encoder achieves an average success rate of 48.0\%, outperforming the contrastive learning-based VITaL, which attains 40.5\%. This performance gap suggests that the reconstruction-based pretraining strategy adopted in the UniVTAC Encoder captures more actionable and physically grounded contact information than contrastive representations alone, which primarily emphasize global feature alignment.

We further conduct ablation studies to analyze the contribution of different tactile perception pathways in the UniVTAC Encoder. As shown in Table~\ref{tab:ablation-pathway}, we evaluate several variants of our method under the same ACT-based policy framework, systematically ablating the tactile input modalities: contact pathway only(\textit{Contact}), shape pathway only(\textit{Shape}), and their combination (\textit{Contact+Shape}), with the from-scratch(\textit{Scratch}) setting serving as the baseline that uses no pretraining and trains the tactile encoder from random initialization.

\begin{table}[t]
\centering
\setlength{\tabcolsep}{4pt} 
\caption{\textbf{Pathways Ablation.} We evaluate the contribution of different tactile perception pathways in the UniVTAC Encoder under the ACT-based policy framework, including contact-only, shape-only, and their combination, with a from-scratch baseline for comparison. (\textbf{Bold}: the best results; \underline{Underlined}: the second-best results).}
\label{tab:ablation-pathway}
\begin{tabular}{l|ccccc}
\toprule
\rowcolor{gray!10}
\textbf{Task} & \textbf{Scratch} & \textbf{Contact} & \textbf{Shape} & \textbf{Contact+Shape} & \textbf{Full} \\
\midrule
Lift Bottle        & 56.0 & 59.0 & 51.0 & \underline{62.0} & \textbf{71.0} \\
Pull-out Key      & 40.0 & \underline{42.0} & 35.0 & 28.0 & \textbf{46.0} \\
Lift Can           & \textbf{33.0} & 12.0 & 26.0 & 18.0 & \underline{29.0} \\
Put Bottle in Shelf & \underline{34.0} & 24.0 & 25.0 & \textbf{39.0} & 31.0 \\
Insert Hole        & 3.0  & \textbf{32.0} & 20.0 & 21.0 & \underline{24.0} \\
Insert HDMI        & 16.0 & 24.0 & \underline{25.0} & 23.0 & \textbf{28.0} \\
Insert Tube        & 45.0 & 43.0 & 53.0 & \textbf{60.0} & \underline{56.0} \\
Grasp Classify     & 45.0 & \underline{99.0} & 98.0 & \textbf{100.0} & \underline{99.0} \\
\midrule
\textbf{\textit{Average}}    & 34.0 & 41.9 & 41.6 & \underline{43.9} & \textbf{48.0} \\
\bottomrule
\end{tabular}
\end{table}

Our full model, which integrates geometric shape, contact force and relative pose through the proposed UniVTAC pretraining scheme, achieves an average success rate of 48.0\%, outperforming all ablated variants. Notably, the performance gain is most pronounced in fine manipulation tasks requiring precise tactile feedback. The \textit{Contact+Shape} variant attains the second-highest average (43.9\%), yet still underperforms the full model, suggesting that the pose perception is essential, especially in the tasks needing pose awareness like \textit{Lift Bottle}.

Importantly, even individual tactile pathways yield significant gains over training from scratch: \textit{Contact} (+7.9\%) and \textit{Shape} (+7.6\%) demonstrate consistent improvements across most tasks, validating the effectiveness of each design component.

These results collectively validate that geometric shapes, contact dynamics and pose perception are essential for dexterous manipulation and joint optimization of the full pipeline yields superior performance compared to modular or partial usage of tactile information.

\subsection{Tactile Data Scaling for Encoder Pretraining}

\begin{figure}[t]
    \centering
    \includegraphics[width=0.48\textwidth]{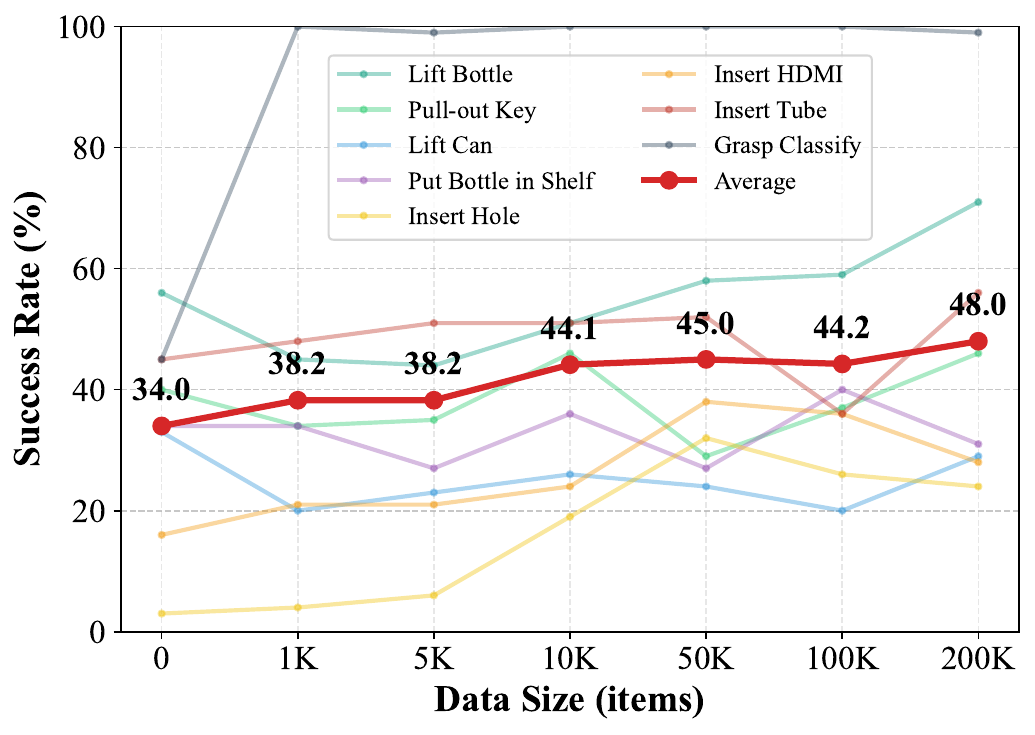}
    \caption{\textbf{Impact of Pretraining Data Scale on Encoder Effectiveness}. Downstream policy performance improves consistently with increasing amounts of synthetic tactile data, highlighting the benefits of large-scale simulated experiences for representation learning.}
    \label{fig:scaling_results}
\end{figure}

\begin{table}[t]
\centering
\setlength{\tabcolsep}{2.5pt} 
\caption{\textbf{Per-Task Success Rates Across Pretraining Data Scales}. Performance varies across tasks, with improvements most pronounced in dexterous manipulation at higher data volumes. (\textbf{Bold}: the best results; \underline{Underlined}: the second-best results).}
\label{tab:encoder-scaling}
\begin{tabular}{l|ccccccc}
\toprule
\rowcolor{gray!10}
\textbf{Task} & \textbf{0} & \textbf{1K} & \textbf{5K} & \textbf{10K} & \textbf{50K} & \textbf{100K} & \textbf{200K}  \\
\midrule
Lift Bottle        & 56.0 & 45.0 & 44.0 & 51.0 & 58.0 & \underline{59.0} & \textbf{71.0} \\
Pull-out Key      & \underline{40.0} & 34.0 & 35.0 & \textbf{46.0} & 29.0 & 37.0 & \textbf{46.0} \\
Lift Can & \textbf{33.0} & 20.0 & 23.0 & 26.0 & 24.0 & 20.0 & \underline{29.0} \\
Put Bottle in Shelf & 34.0 & 34.0 & 27.0 & \underline{36.0} & 27.0 & \textbf{40.0} & 31.0 \\
Insert Hole        & 3.0  & 4.0  & 6.0  & 19.0 & \textbf{32.0} & \underline{26.0} & 24.0 \\
Insert HDMI        & 16.0 & 21.0 & 21.0 & 24.0 & \textbf{38.0} & \underline{36.0} & 28.0 \\
Insert Tube        & 45.0 & 48.0 & 51.0 & 51.0 & \underline{52.0} & 36.0 & \textbf{56.0} \\
Grasp Classify     & 45.0 & \textbf{100.0} & \underline{99.0} & \textbf{100.0} & \textbf{100.0} & \textbf{100.0} & \underline{99.0} \\
\midrule
\textbf{\textit{Average}}    & 34.2 & 38.3 & 38.3 & 44.1 & \underline{45.0} & 44.3 & \textbf{48.0} \\
\bottomrule
\end{tabular}
\end{table}

To investigate the effect of the amount of simulation-synthesized tactile manipulation data on encoder performance, we conduct a data-scaling study for tactile pretraining. Specifically, we pretrain the encoder using different volumes of synthetic tactile data, including 0, 1k, 5k, 10k, 50k, 100k, and 200k samples, and then evaluate the resulting models after pretraining. Detailed quantitative results are reported in Table~\ref{tab:encoder-scaling}, and the trend of average success rate with respect to data scale is shown in Figure~\ref{fig:scaling_results}.

The results indicate that model performance increases monotonically as the amount of simulation data grows, and a clear scaling trend can be observed. This suggests that expanding the diversity and scale of tactile manipulation data consistently improves the effectiveness of the pretrained encoder and its contribution to downstream policy performance.

\subsection{Real-World Experiments}

\begin{figure*}[h]
    \centering
    \includegraphics[width=1.0\textwidth]{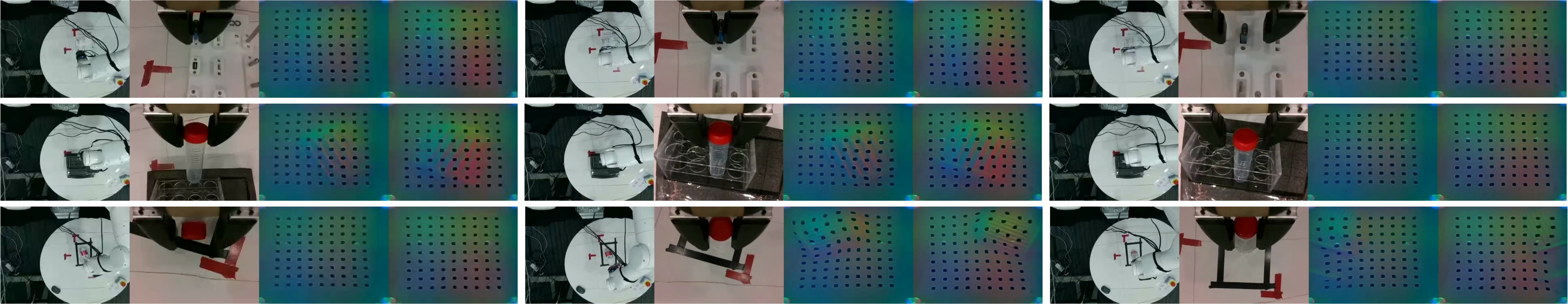}
    \caption{\textbf{Real-world Task Key Frames}.
Representative key frames from three real-world visuo-tactile manipulation tasks, showing synchronized wrist RGB images (left) and marker-based tactile observations (right) at the initial approach, contact-rich interaction, and final completion stages. The intermediate frames highlight evolving contact states and deformation cues that support fine-grained alignment and correction beyond vision-only perception.}
    \label{fig:real_world_key_frame}
\end{figure*}

\begin{figure}[h]
    \centering
    \includegraphics[width=0.49\textwidth]{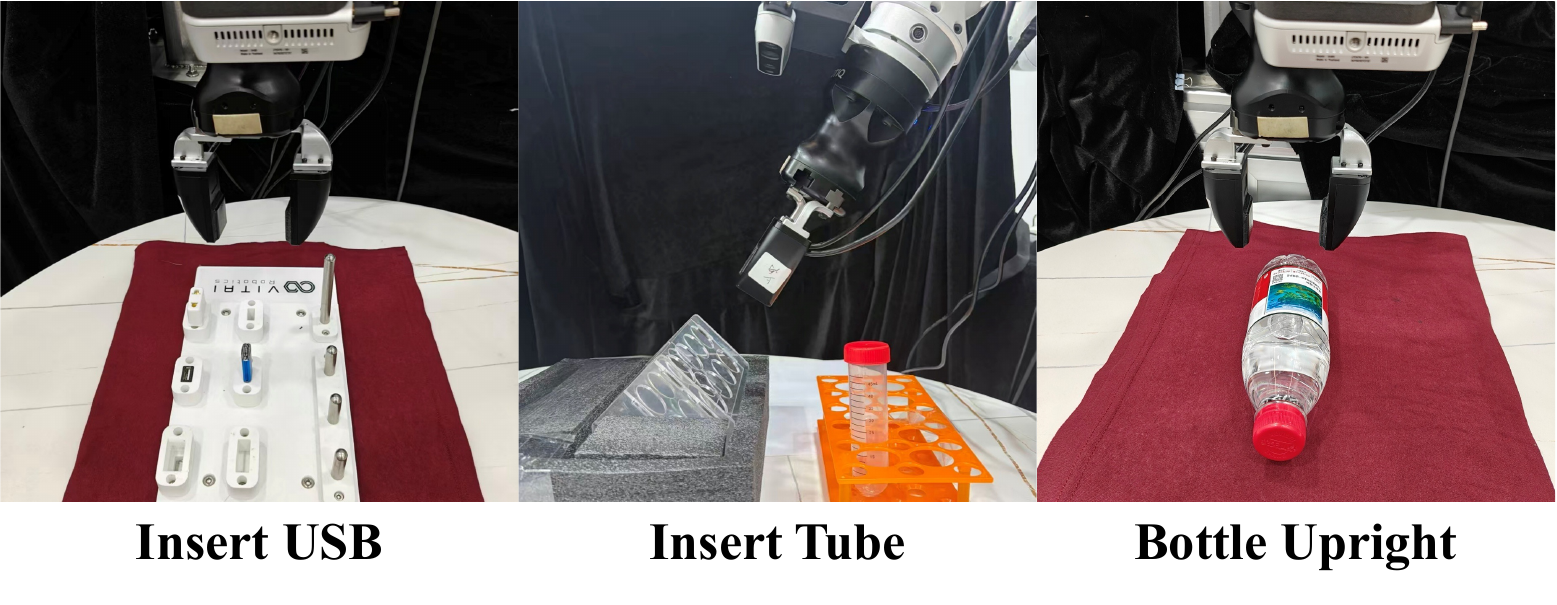}
    \caption{\textbf{Real World Taks}. We illustrate the experimental settings for the three real-world manipulation tasks evaluated in this work.}
    \label{fig:real_world_tasks}
\end{figure}

We evaluate our approach on three physically challenging dexterous manipulation tasks using a real-world robotic platform: \textit{Insert Tube}, \textit{Insert USB}, and \textit{Bottle Upright}. All experiments are conducted on the Tianji Robotics Marvin manipulator, a 7-DoF robotic arm equipped with a parallel gripper and a wrist-mounted RGB camera for visual feedback. Tactile sensing is provided by two ViTai GF225 visuo-tactile sensors embedded in the gripper fingertips, each capturing high-resolution deformation images at 30~Hz. The sensors output marker-based RGB observations that encode rich contact geometry and pressure distribution during interaction. The experimental setups are illustrated in Figure~\ref{fig:real_world_tasks}, and representative key frames of the tasks are shown in Figure~\ref{fig:real_world_key_frame}.

To collect expert demonstrations, we employ VR-based teleoperation using Meta Quest headsets. Human operators control the end-effector pose via hand-held controllers while observing live video streams from the wrist-mounted camera. For each task, we collect 150 high-quality demonstration trajectories, which are used to train an ACT integrated with a diffusion-based policy head. The policy maps sequences of multimodal observations to future actions and outputs joint positions as direct action commands.

The UniVTAC Encoder is first pretrained in simulation using large-scale synthesized visuo-tactile contact data to learn generalizable contact representations. To rigorously evaluate the generalization of the pretrained feature, the encoder is transferred to the real robot without further fine-tuning at deployment time. We conduct ablation studies comparing policies with tactile input enabled via the UniVTAC Encoder against vision-only baselines. All policies are trained separately for each task and evaluated over 20 rollouts under identical initial conditions, with task success determined by human observers.

\begin{table}[t]
\centering
\caption{\textbf{Real-World Evaluation Results.} We report the real-world performance on three manipulation tasks, comparing a vision-only observation setting with a vision-and-tactile fusion setting.}
\label{tab:real_world}
\small
\setlength{\tabcolsep}{10pt}
\renewcommand{\arraystretch}{1.2}
\begin{tabular}{l|cc}
\toprule
\rowcolor{gray!10}
\textbf{Task} & \textbf{Vision} & \textbf{Vision + UniVTAC} \\
\midrule
Insert Tube    & 55.0   & \textbf{85.0}          \\
Insert USB     & 15.0  & \textbf{25.0}       \\
Bottle Upright & 60.0     & \textbf{95.0}     \\
\midrule
\textbf{\textit{Average}} & 43.3 & \textbf{68.3} \\
\bottomrule
\end{tabular}
\end{table}

We further analyze the real-world experimental results to understand how the UniVTAC Encoder contributes to performance gains across different manipulation tasks. As summarized in Table~\ref{tab:real_world}, integrating the UniVTAC Encoder leads to consistent improvements across all evaluated tasks, with an average success rate increase of 25\%. In fine-grained insertion tasks, including \textit{Insert Tube} and \textit{Insert USB}, policies augmented with the UniVTAC Encoder achieve performance gains of 30\% and 10\%, respectively. These improvements indicate that UniVTAC provides high-quality tactile feedback that enables policies to reason over subtle contact cues during precise alignment and insertion. Successful execution in these tasks often depends on detecting partial misalignment and responding with incremental corrective motions, which cannot be reliably inferred from vision alone.

In the \textit{Bottle Upright} task, incorporating the UniVTAC Encoder results in a performance improvement of 35\%. This outcome highlights that the benefits of UniVTAC extend beyond direct contact detection to object pose understanding. Although the encoder is pretrained purely in simulation, its representations implicitly capture object orientation and pose-related cues that are critical for maintaining stability during manipulation. The strong performance observed in real-world execution demonstrates that such pose-aware representations transfer effectively to physical robotic systems.

Overall, the average improvement of 25\% in real-world success rates demonstrates that the UniVTAC Encoder not only provides informative tactile representations, but also enables effective deployment in physical robotic systems despite being pretrained exclusively on simulation-synthesized data. These results underscore the value of simulation-based visuo-tactile data synthesis and highlight UniVTAC as a practical and scalable approach to enhance real-world dexterous manipulation through tactile perception.

\section{Conclusion}
\label{sec:conclusion}

We presented UniVTAC, a unified simulation platform for scalable visuo-tactile data generation, representation learning, and benchmarking in contact-rich robotic manipulation. Built on high-fidelity tactile simulation, UniVTAC enables controllable synthesis of diverse visuo-tactile interactions and provides structured supervisory signals for learning tactile-centric representations. Leveraging this platform, we introduced the UniVTAC Encoder, a visuo-tactile encoder pretrained via multi-pathway supervision that captures object shape, contact deformation, and pose information from tactile observations.
To facilitate systematic evaluation, we further proposed the UniVTAC Benchmark, a benchmark comprising eight representative visuo-tactile manipulation tasks that emphasize tactile-dependent reasoning under occlusion and contact uncertainty. Extensive simulation experiments demonstrate that the UniVTAC Encoder consistently improves manipulation performance when integrated with modern policy architectures, while real-world evaluations confirm effective sim-to-real transfer despite being pretrained purely in simulation.
We believe UniVTAC establishes a practical and extensible foundation for visuo-tactile manipulation research, enabling scalable data generation, robust tactile representation learning and fair benchmarking. Future work will extend this framework to more diverse sensor modalities, dynamic interactions, and open-world manipulation scenarios.

\clearpage
%% Use plainnat to work nicely with natbib. 
\bibliographystyle{plainnat}
\bibliography{references}

\clearpage
\appendices

\section{Task Descriptions for the UniVTAC Benchmark}
\label{task-description}

The tasks included in the UniVTAC benchmark are described in Table \ref{tab:benchmark_description}.

\begin{table}[htbp]
    \centering
    \caption{\textbf{Task Descriptions for the UniVTAC Benchmark.}}
    \begin{tabular}{>{\centering\arraybackslash}m{1.2cm} | m{6cm}} 
    \toprule
    \rowcolor{gray!10}
    \textbf{Task}   & \textbf{Description}        \\ \midrule 
    \textit{Lift Bottle} & A bottle rests on the ground plane facing a vertical wall. The robot grasps the bottle and lifts it vertically, keeping its final base within 5 cm of the wall. \\ \midrule 
    \textit{Pull-out Key} & A key sits in a slot with random initial rotation. The robot rotates the key until sensing mechanical resistance, then pulls it straight out. \\ \midrule
    \textit{Lift Can} & One of three cylindrical cans (4, 5, and 6 cm diameter) lies horizontally on the ground plane. The robot grasps the can and lifts it vertically without slippage. \\ \midrule
    \textit{Put Bottle in Shelf} & A bottle stands upright before a shelf. The robot grasps the bottle, then positions it into the shelf cavity. \\ \midrule
    \textit{Insert Hole} & An cube with inclined hole (60° or 120° orientation) lies on the ground plane. The robot explores the hole's geometry through contact, determines its orientation, and inserts a test tube into the hole.\\ \midrule
    \textit{Insert HDMI} & A HDMI connector held by the robot has random rotational offset. The robot aligns and inserts the connector into a fixed slot under rotational uncertainty. \\ \midrule
    \textit{Insert Tube} & A narrow hole (2.05 cm diameter) appears on an inclined surface. The robot inserts a 2.0 cm diameter test tube from a near-center starting position through the tight clearance. \\ \midrule
    \textit{Grasp Classify} & Two cylindrical objects share similar visual appearance but differ in surface texture. The robot first performs tactile contact to perceive texture, then classifies each cylinder and places it at the pre-specified goal region for its class. \\
    \bottomrule
    \end{tabular}
    \label{tab:benchmark_description}
\end{table}

\section{Implementation Details for Simulation Experiments}
\label{implementation-details}

For each task, we collected 50 episodes of synthetic data used for training. All ACT models were trained for a total of 4,000 optimization steps using a batch size of 64. The learning rate was set to $1 \times 10^{-5}$ for both the vision and tactile encoders, with a weight decay of $1 \times 10^{-4}$. 

The model employs a transformer-based architecture with 4 layers in the encoder and 7 layers in the decoder. For positional encoding, we adopt fixed sine-cosine embeddings for visual features, while learnable positional embeddings are used for tactile features. The model predicts a sequence of 50 future actions from the current observation, which are executed using time aggregation to produce smoother and more stable motor outputs.

Visual observations are primarily captured from a third-person camera view. However, task-specific configurations vary: both \textit{insert tube} and \textit{lift bottle} utilize multi-view inputs, combining third-person and wrist-mounted camera views; all other tasks use only the third-person view.

\section{Details for Real-world Experiments}
\label{realworld-details}

\begin{figure}[h]
    \centering
    \includegraphics[width=0.9\linewidth]{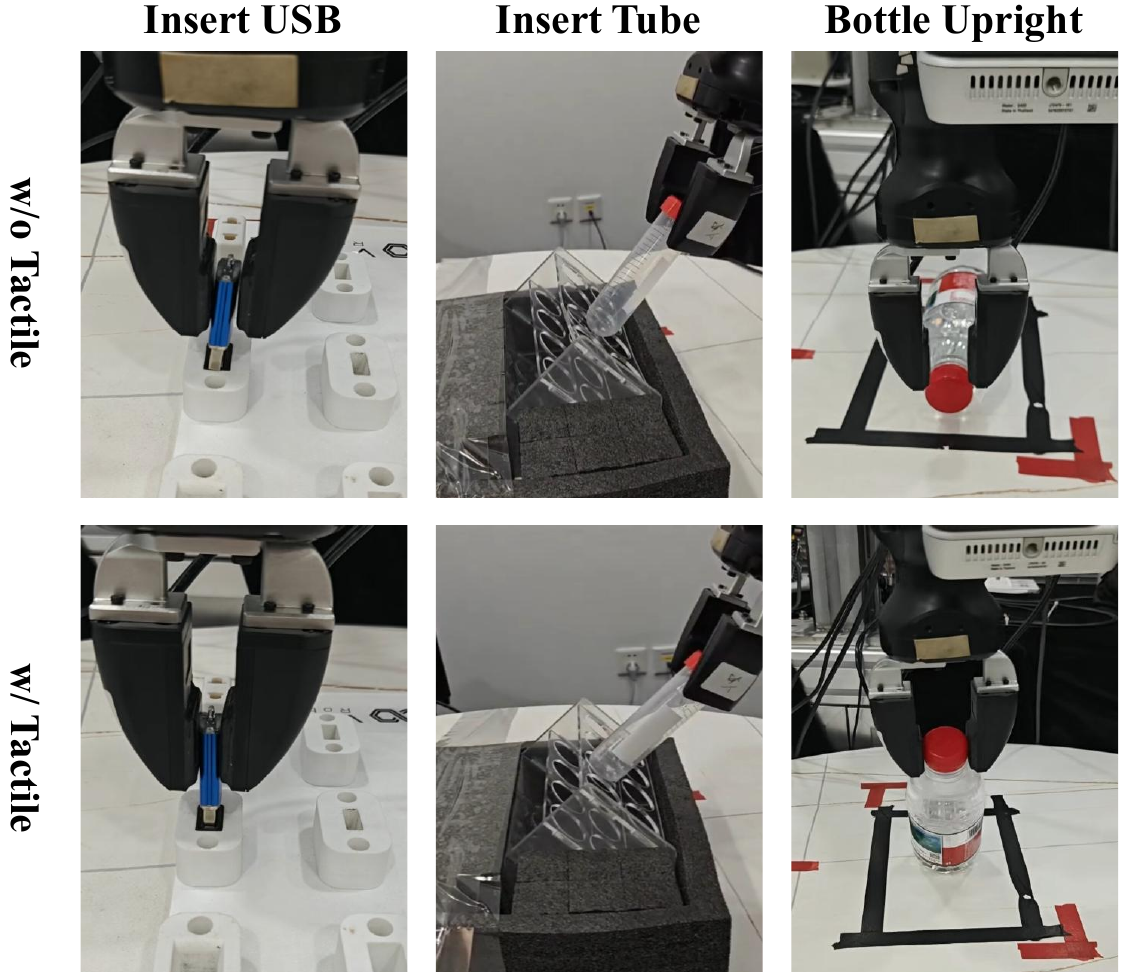}
    \caption{\textbf{Rollout Key Frames.} Tactile feedback enables smoother, more adaptive insertion behaviors with reduced collisions.}
    \label{fig:rollout_key_frame}
\end{figure}

We provide qualitative insights into the role of tactile sensing in real-world task execution. Representative rollout key frames are shown in Figure~\ref{fig:rollout_key_frame}.

In \textit{insert USB}, the tactile-enabled policy uses contact feedback to adjust alignment after initial touch, reducing misinsertion. Without tactile input, the policy often applies excessive force and fails to correct orientation, leading to jamming.

In \textit{insert tube}, the tactile-augmented policy performs gentle probing and real-time correction under tight clearance, avoiding damage to the fixture. In contrast, the vision-only policy tends to push harder when misaligned, causing base shift or hole deformation. Recovery is possible for moderate errors with tactile, but fails under large offsets in both settings.

For \textit{upright bottle}, tactile feedback improves grip stability and motion smoothness. The vision-only policy exhibits noticeable jitter during lifting, likely due to delayed visual feedback and lack of slip detection.

Overall, tactile sensing enhances behavioral compliance and reduces destructive interactions, which are critical for safe and reliable deployment on real world.
\end{document}